\documentclass[10pt]{article} 

\usepackage[accepted]{rlj} 

\usepackage{amssymb}            
\usepackage{mathtools}          
\usepackage{mathrsfs}           
\usepackage{graphicx}           
\usepackage{subcaption}         
\usepackage[space]{grffile}     
\usepackage{url}                
\usepackage{lipsum}             

\usepackage{booktabs}
\usepackage{bbm}
\usepackage[none]{hyphenat}     

\newcommand{\defi}{\overset{\text{def}}{=}}

\title{Toward Agents That Reason About Their Computation}

\setrunningtitle{Toward Agents That Reason About Their Computation}

\author{\mbox{Adrian~Orenstein\textsuperscript{1,2}},
\mbox{Jessica~Jiajing~Chen\textsuperscript{1,2,3}},
\mbox{Gwyneth~Anne~Delos~Santos\textsuperscript{1,2}},
\mbox{Bayley~Sapara\textsuperscript{1,2}},
\mbox{Michael~Bowling\textsuperscript{1,2,4}}}

\emails{aorenste@ualberta.ca}

\affiliations{
$^{1}$\textbf{Department of Computing Science, University of Alberta, Canada}\\
$^{2}$\textbf{Alberta Machine Intelligence Institute (Amii)}\\
$^{3}$\textbf{Department of Computer Science, University of Toronto, Canada}\\
$^{4}$\textbf{CIFAR AI Chair}
}

\contribution{

    We show that agents with actions that change their compute cost are able to learn to adapt their compute at run time in the Arcade Learning Environment (ALE). 
    We observe an improvement in performance on 75\% of games and a 3 times improvement in efficiency averaged across games. 
    We support claims in a performance and efficiency improvement in Section~\ref{sec:performance_and_cost} and that this is changing throughout the lifetime of the agent in Section~\ref{sec:learned_from_experience}.
    }
    {
    The significance in this result is in seeing compute use and performance can co-evolve throughout learning without necessarily incurring an average drop in performance.
    This is a result to contrast that much of the agent's design is defined by people, and fixed after design time, leaving little control for the agent to adapt its computation.
    }

\contribution{
    We show that agents learn compute efficient behaviors that are tailored to each game and adapt to the changing circumstances within a game. 
    We support this claim using the videos included in the supplementary material, and the three investigations in  Section~\ref{sec:agent_strategies}.
    }
    {
    These videos show the agent changes its decision rate on a per-game basis, and adapts to moments in each game.
    }

\contribution{
    We show that agents are responsive to the experienced compute cost given as a penalty to their reward. 
    We support this claim in Section~\ref{sec:agents-respond-to-compute-costs}.
}
{
    As we are encouraging agent designs to incorporate actions that change compute use and feedback on compute use, it is important to show how agents respond more generally to different trade-offs, as we would like agents to be sensitive to their problem specification.
}


\summary{

While reinforcement learning agents can achieve superhuman performance in many complex tasks, they typically do not become more computationally efficient as they improve. 
In contrast, humans gradually require less cognitive effort as they become more proficient at a task.
If agents could reason about their compute as they learn, could they similarly reduce their computation footprint? 
Agents \textit{reason} by explicitly representing the expected value of selecting an action and how it impacts the expected return.
And so, if agent could learn the expected value over both actions and how they use compute, we could have more energy efficient agents or free up compute cycles for other processes like planning.
In this paper, we experiment with showing agents the cost of their computation and giving them the ability to control when they use compute.
An interactive version of this paper is available at \href{https://adrianorenstein.com/paper/toward-compute}{\texttt{adrianorenstein.com/paper/toward-compute}}.
}

\begin{document}

\makeCover  
\maketitle  

\begin{abstract}
While reinforcement learning agents can achieve superhuman performance in many complex tasks, they typically do not become more computationally efficient as they improve.
In contrast, humans gradually require less cognitive effort as they become more proficient at a task.
If agents could reason about their compute as they learn, could they similarly reduce their computation footprint?
If they could, we could have more energy efficient agents or free up compute cycles for other processes like planning.
In this paper, we experiment with showing agents the cost of their computation and giving them the ability to control when they use compute.
We conduct our experiments on the Arcade Learning Environment, and our results demonstrate that with the same training compute budget, agents that reason about their compute perform better on 75\% of games.
Furthermore, these agents use $3$ times less compute on average.
We analyze individual games and show where agents gain these efficiencies.
An interactive version of this paper is available at \href{https://adrianorenstein.com/paper/toward-compute}{\texttt{adrianorenstein.com/paper/toward-compute}}.

\end{abstract}


\section{Introduction}

As reinforcement learning (RL) agents get better at a task, they typically do not become more energy efficient with experience.
This is unlike humans, who adapt their cognitive process over time and can learn to devote cognitive effort with intention~\citep{millerIntegrativeTheoryPrefrontal2001}.
Agents, by contrast, often have their \textit{computational processes} (a.k.a. \textit{compute}) fixed;
their processes for sensing, acting, learning and planning do not change after design time.
So, when the problem demands more computation in some moments and less in others, the agent's computational process does not adapt.

There are many ways that agents are designed to use compute, and these choices have mostly been found by agent designers~\citep[e.g.,][]{mnih_dqn, silver_go, vinyals_alphastar, marc_balloon, peter_gtsophy, degrave_fusion, mankowitz_alphadev}.
Designers shape how and with what frequency the agent should use compute so that it can more effectively scale with available resources.
As one simple example, the Arcade Learning Environment~\citep{bellemare_ale} (ALE) outputs 60 frames per second.
Instead of making agents process each frame, making the agent process only every fifth frame as is the default with the ALE speeds up the execution by five times, with a negligible reduction in performance~\citep{bellemare_early_atari}.

If agents could control their compute and received feedback on how they used it, they could begin to make these choices for themselves.
But at present, designers choose how compute is used --- designers decide when the agent should act or learn or plan, and the ``cost'' of compute is not made available to the agent.
Additionally, agents are not typically long lived~\citep{suttonAlbertaPlanAI2022}, they are created to use compute in a certain way and replaced once their design limits them.
This creates a \textit{cycle of design}: where an agent is created, their performance is observed, the design is evaluated, and a new agent is improved that fits the trade-off known at design time.
This cycle of design suggests that agents could do the same for themselves if they had access to similar actions and received feedback about their compute costs.

This work represents a foray into agents that can reason about their own computational processes.
Since agents are capable of learning to make effective choices within the RL formalism, we will use the very same RL formalism to frame the challenge of reasoning about compute.
We provide the agent with additional actions that do not have direct external effects, but rather change the agent's usage of compute.
Additionally, we provide feedback that makes the cost of compute explicit alongside traditional environment reward.
We focus on providing the agent with actions that change the rate at which observations are processed and environment actions are computed, while modifying its rewards to reflect the cost of computation.
Each decision to act invokes the agent's whole computational process: constructing state, producing an action, storing experience, and learning.
The agent's decision rate is therefore our proxy for its computation time.
We will show that in this setting agents can effectively learn to adapt their computational processes at run time from just their experience.

Our contributions are as follows:
\begin{enumerate}[leftmargin=1.5em, itemsep=0.2em, topsep=0em]
    \item We show that an agent with the ability to control its compute and see the consequences of its computation can learn to adapt its compute at run time.

    \item These agents perform better on 75\% of the Arcade Learning Environment games, while using the same training compute budget.
          They learn compute efficient behaviors per-game that are on average $3$ times more efficient.
          We show that efficiencies are learned and tailored to each game, and adapt to the changing circumstances within a game.

    \item We show that agents are responsive to the experienced compute cost, showing that an agent's computational processes adapt to the RL task posed to it by the designer.
\end{enumerate}

\section{Reasoning about computation}

In this section, we show that we can use the RL problem formulation to enable agents to reason about their compute.
Our control problem consists of observations and actions.
The agent processes an observation into a state, $S_t$, at a time step, $t$, and computes an action, $A_t$, in the same time step.
This action is then sent to the environment, and in response, a new observation is perceived in the next time step resulting in the next state $S_{t+1}$.
A special part of the agent's observation is a scalar reward signal, $R_{t+1}$, received at every time step.
A discount factor, $\gamma \in [0, 1)$, weights future rewards.
This cycle of interaction repeats in a synchronized manner, and the objective of the RL agent is to maximize the expected episodic discounted return: $G = \mathbb{E}\left[\sum_{k=1}^{T-1} \gamma^{k-1} R_{k+1}\right]$.

\subsection{Framing the objective}

The agent's goals are specified in the reward signal~\citep{suttonRewardHypothesis2004, bowlingSettlingRewardHypothesis2023}, and so, the scalar reward should also provide the consequences of the agent's computational processes.
For instance, consider how an agent may learn when it is important to act while operating a solar panel tracking station.
The solar panel needs to be continually oriented toward the sun to capture sunlight throughout the day, where a better agent is one that maximizes the power exported to the grid.
The agent can learn to track the sun accurately by actuating the panel and observing the exported energy change in value.
However, the agent's computation consumes energy as well, which is energy not exported to the grid.
If the objective incorporated the energy that has been gathered, $r^{\text{task}}_t$, and the compute cost of the agent, $c_t$, a scalar reward can be constructed representing the total power exported to the grid: $r_{t} = r^{\text{task}}_{t} - c_{t}$.
This frames the problem the designers face --- balancing the agent's compute use with performance on the task, but now allows the agent to see feedback on its compute usage.
In the next section, we show how actions that change the cost can enable the agent to choose when to use compute.

\subsection{Controlling the computational processes}
\label{sec:controlling_compute}

The rate at which agents process observations and take action greatly changes their compute cost.
If the agent processes at half the rate, the agent can halve its compute cost.
This is explicitly true for all agents with fixed rates of environment interactions with which they store experience, learn from experience, or generate experience from a model for training or choosing actions.
When an agent takes an action, it could also decide how long it should keep repeating that action until it processes another observation.
This gives the agent direct control over the rate at which it processes observations and takes actions, and therefore control over its compute usage.
The options framework~\citep{sutton_options} abstractly defines temporally extended courses of action, and so, we use options to model our expanded concept of action to let the agent control when it will process the next observation.

We provide the agent with a set of options ${\cal O} = {\cal A} \times {\cal T}$, where ${\cal T} \subset \{1, 2, \ldots\}$ is a set of possible durations an action can be repeated.
We assume the set of durations are given but these could be discovered by the agent itself~\citep{harb_options_with_deliberation_cost, barreto_option_keyboard}.
While an option is being executed the agent uses a negligible amount of compute.
However, the selection of the option to execute is compute intensive, invoking all of the agent's processes for constructing state, producing an action, storing experience, learning, planning, etc.
We tie computation cost $c_t$ at time $t$ directly to whether an option was selected,
\begin{equation}
    c_{t+1} \defi
    \begin{cases}
        c & \text{if an option is selected, at time $t$, or} \\
        0 & \text{during option execution}.
    \end{cases}
\end{equation}
The agent's policy $\pi : {\cal S} \rightarrow \Delta({\cal O})$, maps the current state $s_t$ to a distribution over the set of options, for which $o_t$ is sampled.
We will focus on value-based RL methods that select an option using an option-value function $Q : {\cal S} \times {\cal O} \rightarrow \mathbb{R}$, which are learned from experience.
The option-value function can be updated as in a semi-MDP~\citep{sutton_options}, which due to the fixed option duration is equivalent to employing a $\tau$-step TD-update for option $o_t = \left<a_t, \tau_t\right>$.
The TD error for selecting this option at time $t$ is then:
\begin{equation}
    \delta_t = \left((r_{t+1} - c)+\gamma r_{t+2}+\ldots+\gamma^{\tau-1} r_{t+\tau} + \gamma^\tau\max_o Q\left(s_{t+\tau}, o\right) - Q(s_t, o_t)\right).
    \label{eqn:td-error}
\end{equation}
Note the introduction of the cost of compute $c$ which is incurred at time $t+1$ after the option was selected while the compute cost $c_{t+2} = \ldots = c_{t+\tau} = 0$ and is dropped.
Additionally, much like other conditional computation methods~\citep{gravesAdaptiveComputationTime2017}, if no cost is provided for using computation, option-values won't differentiate options that have significantly different compute costs.

We have provided the agent with a means to control its compute usage through the selection of $\tau_t$ in option $o_t$, while also providing it with feedback about the cost of compute through $c_t$.
By providing a variety of option durations ${\cal T}$, the agent can learn when precisely timed control is necessary (choosing shorter options), and when compute should be conserved (choosing longer options).
We can now explore the central question of the paper: \textit{Can an agent with the ability to control its compute and see the cost of its computation learn to adapt its compute usage at run time?}

\section{Experimental Setup}

We evaluate on the Arcade Learning Environment (ALE)~\citep{bellemare_ale, machado_ale}, which provides many diverse, stochastic games.
Agents use sticky actions with the default probability and a frame-skip of $5$.
Our baseline is Deep Q-Network (DQN)~\citep{mnih_dqn}, with standard modern improvements: Adam optimization~\citep{kingmaAdamMethodStochastic2015}, activation normalization after each convolutional layer~\citep{ioffeBatchNormalizationAccelerating2015, zhangRootMeanSquare2019}, and gradient clipping at magnitude $1$~\citep{schwarzerBiggerBetterFaster2023, clarkRainbowHighPerformance2024}.
These changes improve stability while preserving the core DQN design.

We measure compute by the number of \textit{decisions}, $d$, where each decision invokes the agent’s full process for constructing state, producing an action, storing experience, and learning.
In our experiments, we will often refer to the number of decisions an agent has made $d_T = \sum_{t=1}^T \mathbbm{1}(c_t \ne 0)$ or the agent's decision rate $d_T/T$, which for ease of presentation we present as decisions per second (Hz), or $60 d_T/T$, as all of our experiments are in Atari games that operate at 60 frames per second.
Standard DQN acts every 5 frames, which fixes its decision rate at 12\,Hz.
Agents that choose to make decisions less frequently, reducing their compute usage, would see a decisions per second less than 12\,Hz.
We deliberately focus on counting decisions rather than wall-clock time, or watt-hours as it is hardware and software independent and is directly correlated with floating-point operations, providing a stable basis for comparison.

Our approach, \textit{Compute DQN}, extends DQN by adding actions that control decision frequency.
We generate options using the Cartesian product of the reduced action set per-game and durations ${\cal T} = \{1,2,4,8\}$: ${\cal O} = {\cal A} \times {\cal T}$.
This lets the agent operate at DQN's fixed rate of 12\,Hz (every 5 frames) or 6\,Hz, 3\,Hz and down to 1.5\,Hz (i.e., acting as rarely as every 40 frames), giving coarse control over its decision rate.
Other than that, Compute DQN operates almost identically to DQN treating its options as actions.

Compute DQN \textit{reasons} about compute by explicitly representing the expected value of different compute-actions that each impact the return under our explicit compute-performance trade-off.
It learns a state-option value function, updating after a fixed number of decisions, by training on tuples of experience from its replay buffer, $\left< s_t, o_t, r_{t:t+\tau} - c, s_{t+\tau}\right>$, where $r_{t:t+\tau} = \sum_{i=1}^{\tau} \gamma^{i-1} r_{t+i}$ is the discounted sum of rewards received over the duration of the option.
Additionally, in calculating its TD-update it discounts the value of the stored next state from its target network by $\gamma^\tau$, to account for the uneven option durations as in Equation~\ref{eqn:td-error}.

Both agents are trained with the same compute budget.
We fix training to 40 million decisions, which matches the compute used by DQN when trained on 200 million frames with a frame skip of 5.
Keeping the number of decisions constant lets us focus on compute efficiency over sample efficiency.
Indeed, agents that reduce their compute usage through their decision rate will cause the emulator to step more, enabling agents to complete more episodes within a training run, possibly gaining more diverse tuples of experience.
Although, as we will discuss in the results section, we do not see this concern substantiated in our results, as improved efficiencies do not correlate with greater performance.

As a good score in the arcade environments varies considerably between games and often by orders of magnitude.
This impacts the magnitude of the accumulated reward observed throughout learning differently in each game and presents a challenge in finding a compute cost appropriate for each game that reliably creates interesting behavior.
We find that setting the compute cost to be the ratio of the accumulated undiscounted return $G_T$\footnote{
    The return $G_T$ is based on clipped rewards.
    As is typical with the ALE, agent rewards are clipped to the range $[-1, 1]$ to handle the high variation in rewards scales across games.
    As such, the cost $c$ is also guaranteed to be in the range $[-1, 1]$.
} achieved by a fully trained DQN agent, divided by the decisions to achieve that performance, to produce a cost $c = \tfrac{G_T}{T}$ that results in interesting behavior.
In an actual application, we would expect it to be more natural to set $c$ by an actual tradeoff between the agent achieving its goals and the cost of its compute to pursue those goals.
However, in settings such as the ALE, we use the DQN agent configured for the highest decision-rate to provide the ``performance above all else'' baseline used to set the performance-compute trade-off that is fixed for a game that compute DQN can try to beat.
However, this approach of tying the compute cost to the agent's average reward could give a knob to emphasize the importance of compute efficiency by doubling or halving this calculated $c$, which we explore later in the results section.
Additionally, setting the compute cost to zero appears to simply burden the agent with additional action values that it needs to accurately estimate, rather than giving the agent feedback on its compute.
All results are averaged over 10 independent runs with different seeds for network initialization and environment stochasticity\footnote{\url{https://github.com/AdrianOrenstein/toward-agents-reasoning-about-compute}}.

\section{Results}

We first present overall results across the Arcade Learning Environment's suite of games, then analyze three games in detail.
Our analysis shows that agents effectively learn to efficiently use compute from experience: their strategies improve with experience, are tailored to each game, and adapt to the changing circumstances within a game.
We further show that agents can adapt to different costs of compute, using more or less compute as $c$ is varied.
These results show that agents not only learn to use compute from experience, but also adjust their decision rates consistently across games, game specific moments, and costs.

In all of our results we will focus on two measures: performance and compute usage.
For performance we report the undiscounted return averaged over the last 100 training episodes across 10 seeds, using game score as reward\footnote{
    We removed three games --- Asteroids, PrivateEye, and MontezumaRevenge --- as DQN achieved less than $10\%$ HNS making relative improvements uninformative.
    Nevertheless, Compute DQN was observed to outperform DQN on MontezumaRevenge and Asteroids and was worse on PrivateEye.
}.
We then normalize this score to produce a Human Normalized Score (HNS) ~\citep{mnih_dqn}, allowing us to more fairly compare across games.
For compute use we report the decisions per second (Hz) rate of the agent relative to the 60 frames per second real-time rate of Atari.
All results are obtained while the agent is learning, and exploration is enabled unless stated otherwise.

\subsection{Performance and compute use} \label{sec:performance_and_cost}

\begin{figure}[t]
    \centering
    \noindent\includegraphics[width=\textwidth]{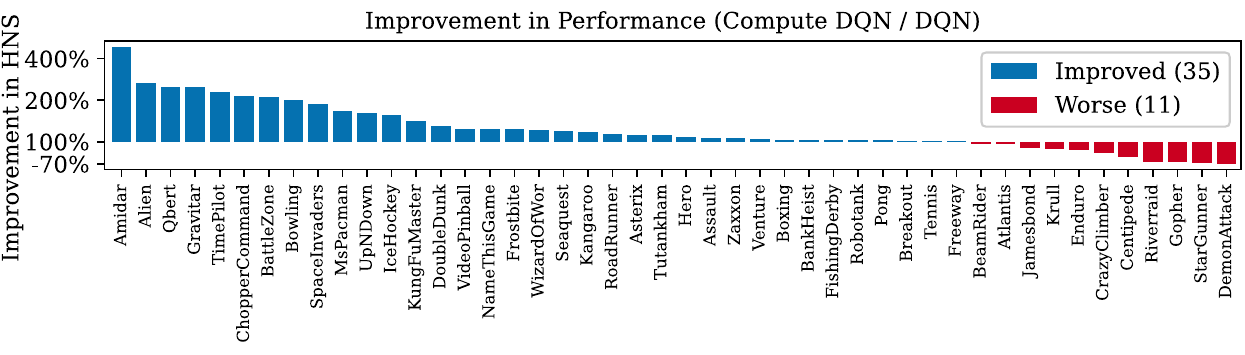}
    \caption{
        Improvement of Compute DQN over DQN in human normalized score (HNS) across 46 Atari games.
        Compute DQN achieves higher HNS in 75\% of games.
        Amidar shows the largest gain of 487\%, while the largest loss is 30\% below DQN.
    }
    \label{fig:arcade_perforamnce}
\end{figure}

Under equal training compute, Compute DQN outperforms DQN on 75\% of Atari games (see Figure~\ref{fig:arcade_perforamnce}).
The largest gain is on Amidar with 487\%, while the largest loss is on DemonAttack with 30\% below DQN (see Figure~\ref{fig:arcade_perforamnce_raw} in the Appendix for raw HNS of the algorithms).
Note that the goal was to give actions to allow the agent to balance the tradeoff between performance and compute, not improve performance.
It may seem odd then that there would be a performance improvement at all.
However, this result appears consistent with others who have suggested repeated actions improve credit assignment~\citep{macro_actions} possibly as the result of using an n-step update;
or repeat actions do better exploration~\citep{dabneyTemporallyextendedEgreedyExploration2020}, or are more sample efficient~\citep{frameskip_as_a_hyper}.
To determine whether a lower decision rate benefits performance, we retrain DQN set to a decision rate closer to what Option DQN learns in Appendix~\ref{sec:dqn_using_learned_avg_decision_rate}, and find that the decision rate alone yields little gains over Compute DQN.
Additionally, to determine whether this improvement is due to improved credit assignment from n-step updates (which options implement), we compare Rainbow and Compute Rainbow in Appendix~\ref{sec:rainbow} and find Compute Rainbow still outperforms Rainbow.
In summary, giving an agent actions and feedback so that it can reason about their compute may not necessarily reduce performance.

\begin{figure}[h]
    \centering
    \noindent\includegraphics[width=\textwidth]{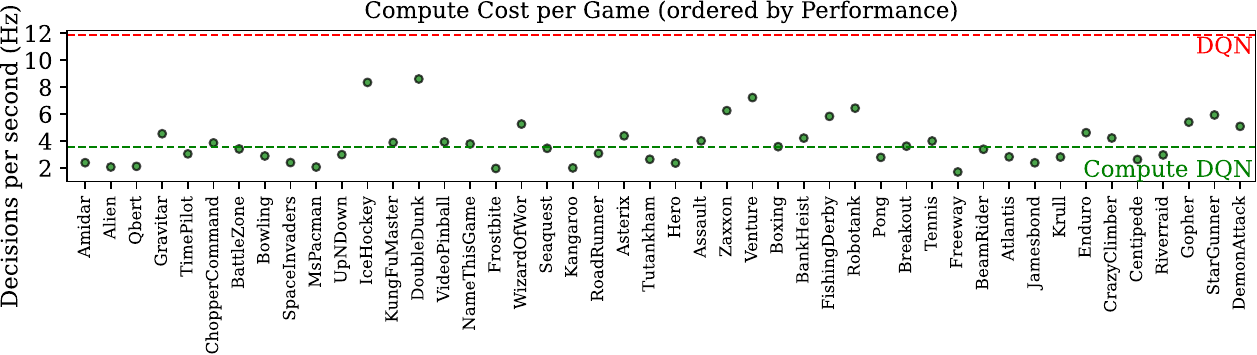}
    \caption{
        Compute DQN learns game-specific decision rates across the Arcade suite.
        Each green dot shows the average decisions per second for a single game, measured over 10 independently trained agents.
        The dashed green line indicates the suite-wide mean of 3.6\,Hz.
        For comparison, Compute DQN uses $3.4$ times less compute than DQN, which acts at a fixed 12\,Hz (red dashed line).
        These results demonstrate that Compute DQN can adapt its compute use per game while maintaining performance.
    }
    \label{fig:overall-decision-rate-per-game}
\end{figure}

Compute DQN's primary goal is to reduce compute usage, for which it is also very effective (see Figure~\ref{fig:overall-decision-rate-per-game}).
On average it uses 3.4 times fewer decisions per second than DQN (3.6\,Hz vs.\ 12\,Hz).
Paying closer attention to each game, games where Compute DQN finds the most efficient behavior (being lower on the graph)  do not necessarily correspond to realizing a benefit from stepping the emulator more frequently over the same training budget.
As such, we observe that performance and efficiency have largely more to do with the game in particular rather than a greater efficiency being predictive of greater performance.





Before moving on to analyzing efficiencies at runtime, we want to draw a comparison between Compute DQN's average decision rate and human decision rates.
While human action rates are not available for the ALE, StarCraft II professionals average $\sim$174 actions/min~\citep{vinyals_alphastar};
Compute DQN’s learned average of 3.6\,Hz ($\approx$216 actions/min) is closer to this human scale than DQN’s fixed 720 actions/min, suggesting that Compute DQN is learning a decision rate similar to what people learn to operate at.
As the arcade games were originally designed for people to play, giving the agent the means to reason about their compute (both actions and feedback) appears to benefit agents as well.
Quite possibly, this may beneficially align representations or strategies more with people, as game design incorporates knowledge that people reason (can act and get feedback) about their cognition as well.
These results raise the question now of how the agent's decision-making is achieved in the first place and whether actions toward performance and compute during play appear deliberate.
We explore this next by taking Castro's~\citeyearpar{castroDefenseAtariALE2024} advice and narrow in on a few games rather than presenting metrics in aggregate.

\subsection{Efficient usage of compute is learned from experience}
\label{sec:learned_from_experience}

\begin{figure}[h]
    \centering
    \noindent\includegraphics[width=\textwidth]{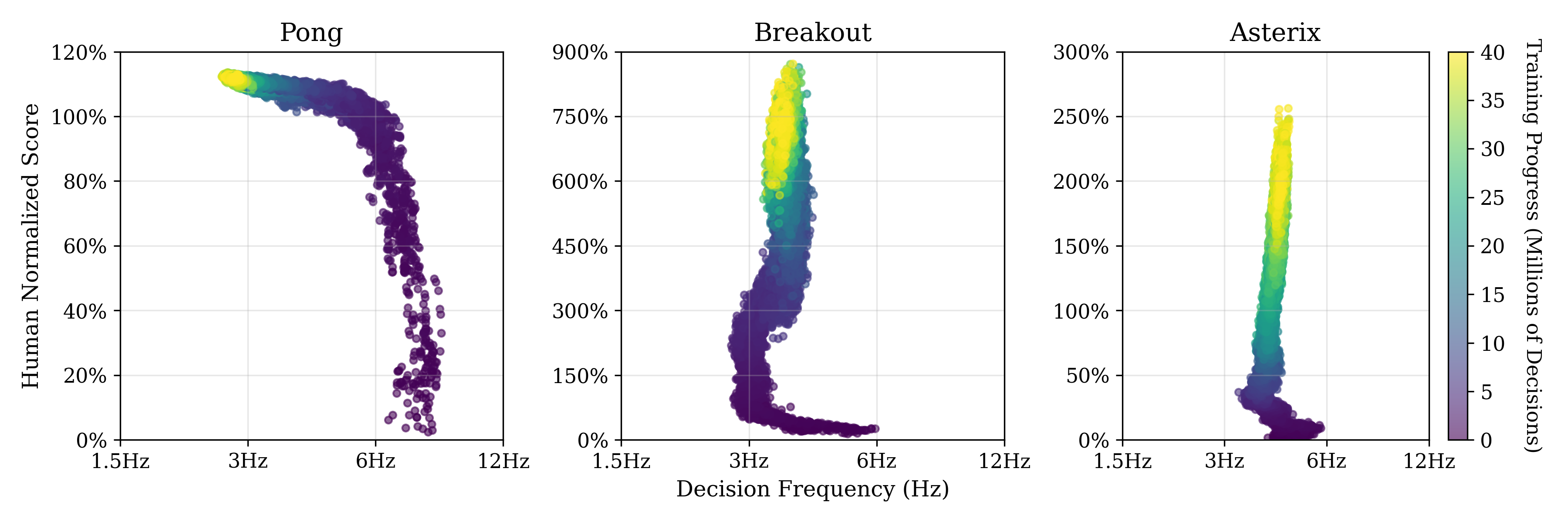}
    \caption{
        Population dynamics of learning: each point is a moment in time during training, reported across 10 seeds per game.
        Color encodes training progress.
    }
    \label{fig:all_games_experience}
\end{figure}

Now we analyze how decision rates and performance co-evolve during training.
First, we can see in Figure~\ref{fig:all_games_experience} that agents indeed learn to adapt their decision rate through trial and error learning;
however, the exact trend varies between the games.
In Pong, performance improves considerably early in learning without a distinct reduction in the decision rate.
Only after the policy is quite strong do we see the agent favoring lower decision rates, finding action sequences that still preserve high scores while reducing the computation cost.
This contrasts with Breakout and Asterix, where both see a sharp drop in the decision rate early in learning.
Before good sequences of actions are found, the optimal choice is to simply make fewer decisions and avoid the computation cost.
This is most clear in Breakout where the decision rate is halved in the first million decisions.
As the agent's policy improves in Breakout and Asterix, though, the tradeoff reverses as the agent begins to return to taking more frequent actions that now lead to considerable increases in performance.

\subsection{Investigating the learned strategies}
\label{sec:agent_strategies}

\begin{figure}[h]
    \centering
    \noindent\includegraphics[width=\textwidth]{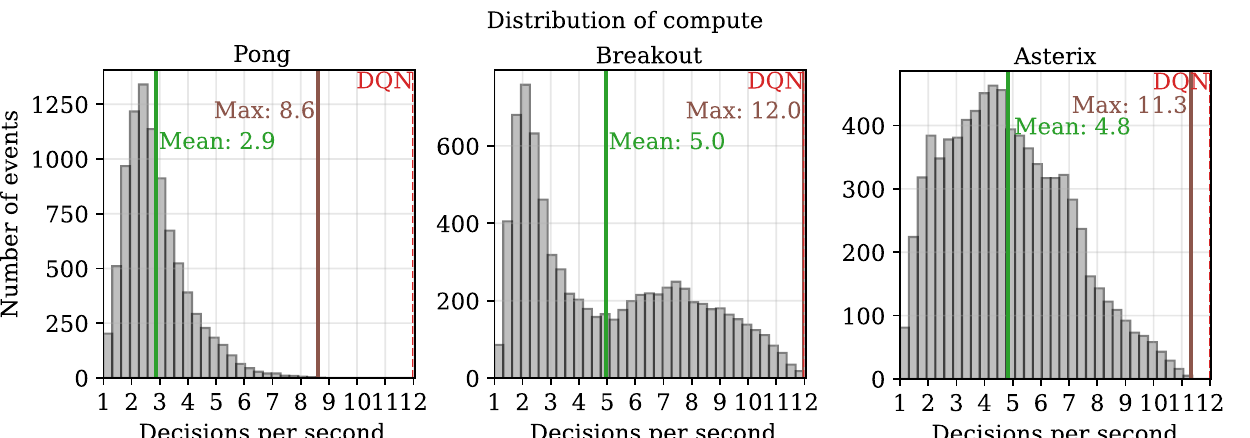}
    \caption{
        The decision rate within a trajectory differs by game.
        Pong and Asterix are uni modal, whereas Breakout is bimodal.
    }
    \label{fig:distribution_of_compute}
\end{figure}

We now examine a single agent trained to completion for each game and consider its learned policy.
Figure~\ref{fig:distribution_of_compute} shows the distribution of the decision rate over a complete trajectory.
The agent's learned decision rate is different across games and shows distinct game-specific patterns.
In Pong, decision rates cluster tightly around the mean, with only occasional spikes.
Breakout shows a bimodal distribution, indicating that the game may alternate between moments that are less intensive and others that require frequent decision-making.
Asterix shows a broader, more uniform distribution.

We further analyze these trajectories individually in the next section.
To let readers form their own view first, we provide videos showing the game, and the agent's decision rate\footnote{Videos on the three discussed games along with others are in the supplementary material.
}.
During the video capture, we take an agent who has completed training, we turn off all training code, and we run the agent deterministically.
We do this setup so that the viewer can attribute actions taken by the agent to the greedy deterministic policy.
The videos and game captures of the agents behavior in this section are the only results in this paper where training and exploration are turned off.

\subsubsection{Pong}

\begin{figure}[h]
    \centering
    \begin{subfigure}[t]{0.48\textwidth}
        \centering
        \includegraphics[width=\textwidth]{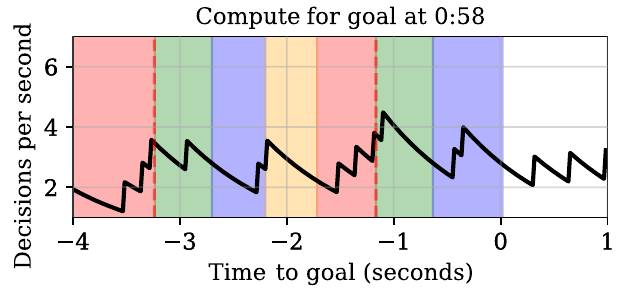}
        \caption{Two hits: one fast, the other is slow.}
        \label{fig:pong-compute-anticipate}
    \end{subfigure}
    \hfill
    \begin{subfigure}[t]{0.48\textwidth}
        \centering
        \includegraphics[width=0.98\textwidth]{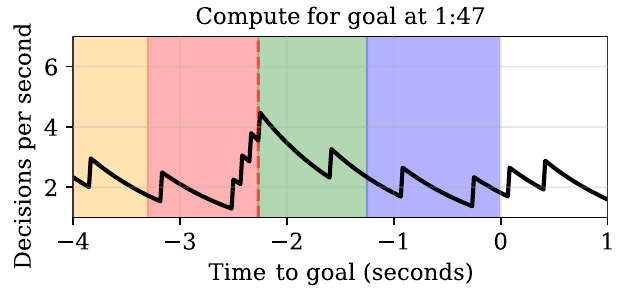}
        \caption{A slow ball hit by the agent.}
        \label{fig:pong-compute-conserve}
    \end{subfigure}
    \caption{
        Exponentially decayed average of decision rate for Compute DQN on segments of an episode in Pong.
        Coloured regions indicate when {\color{orange} the opponent has just hit the ball}, the {\color{red} ball is approaching the player and dotted line indicates when the ball is hit}, {\color{green} the player has just hit the ball}, and {\color{blue} the ball is heading toward the opponent}.
        There are typically rises in decision rate in the red region in anticipation of the ball reaching the player's paddle.
    }
    \label{fig:pong-compute-analysis}
\end{figure}


In Pong, the agent reduces compute when acting would not get the agent in a better position to strike the ball.
Instead, it tends to make more frequent actions moments before hitting the ball (in red), or to reposition.
Additionally, we see general trends of conserving compute, where the decision rate drops after the player hits the ball (in green), and when the ball is heading toward the opponent (in blue).

\subsubsection{Breakout}

\begin{figure}[h]
    \noindent\includegraphics[width=\textwidth]{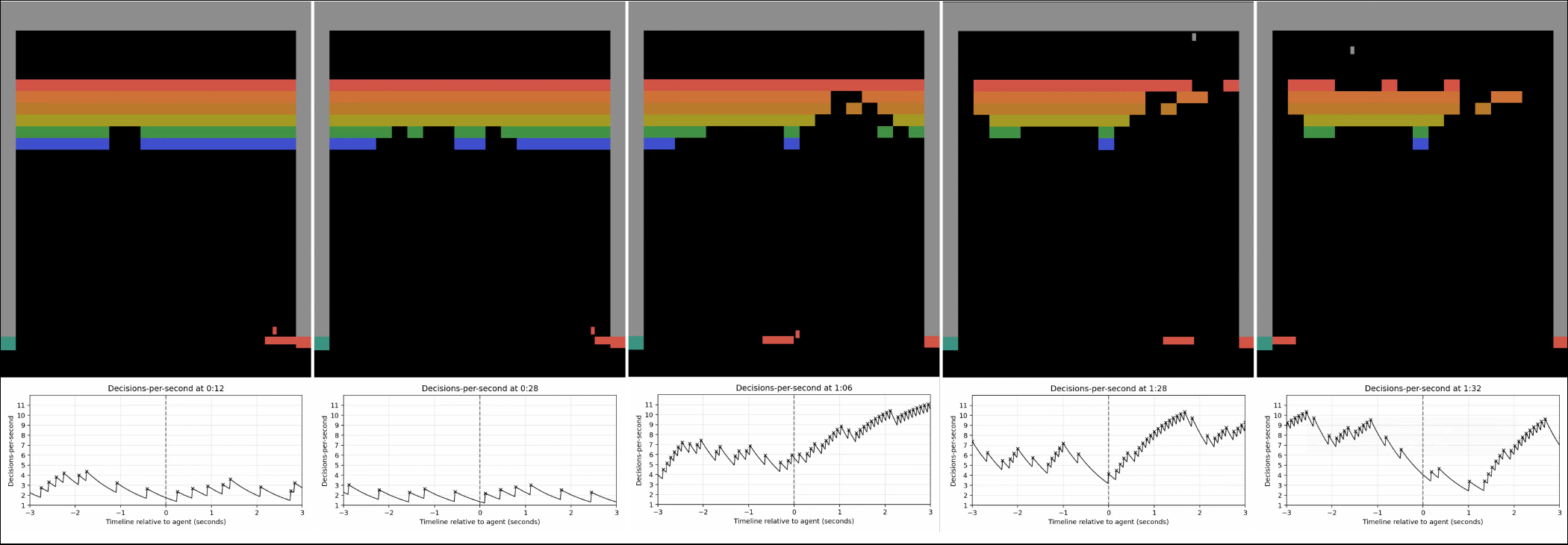}
    \caption{
        A sequence of observations (top) with an associated graph of the exponentially decayed average of decision rate at that moment (bottom) for Compute DQN in Breakout.
        As the game progresses the decision rate increases considerably, in correlation with the ball speeding up as blocks are cleared and a ``breakout'' occurs.
        The final image shows an example decision rate dropping while the ball is behind the wall of blocks.
    }
    \label{fig:decisions-per-second-breakout-longform}
\end{figure}

The difficulty of Breakout increases in well-defined stages: the ball accelerates after  consecutive blocks have been hit, and striking the back wall both increases speed and shrinks the paddle~\citep{BreakoutVideoGame2025}
These mechanics gradually increase the precision of control needed to continue to keep the ball in play and score points.
In Figure~\ref{fig:decisions-per-second-breakout-longform}, we see that the first two images show moments when the ball speed increases by a small amount, and the agent does not increases its use of compute.
Whereas the next two images show two moments: the first where the ball further increases in speed, and the next where the paddle width decreases, both resulting in increases in the decision rate.
We see in the final image that the agent conserves compute when the ball is stuck behind the wall and no immediate action is needed.

\subsubsection{Asterix}

\begin{figure}[h]
    \noindent\includegraphics[width=\textwidth]{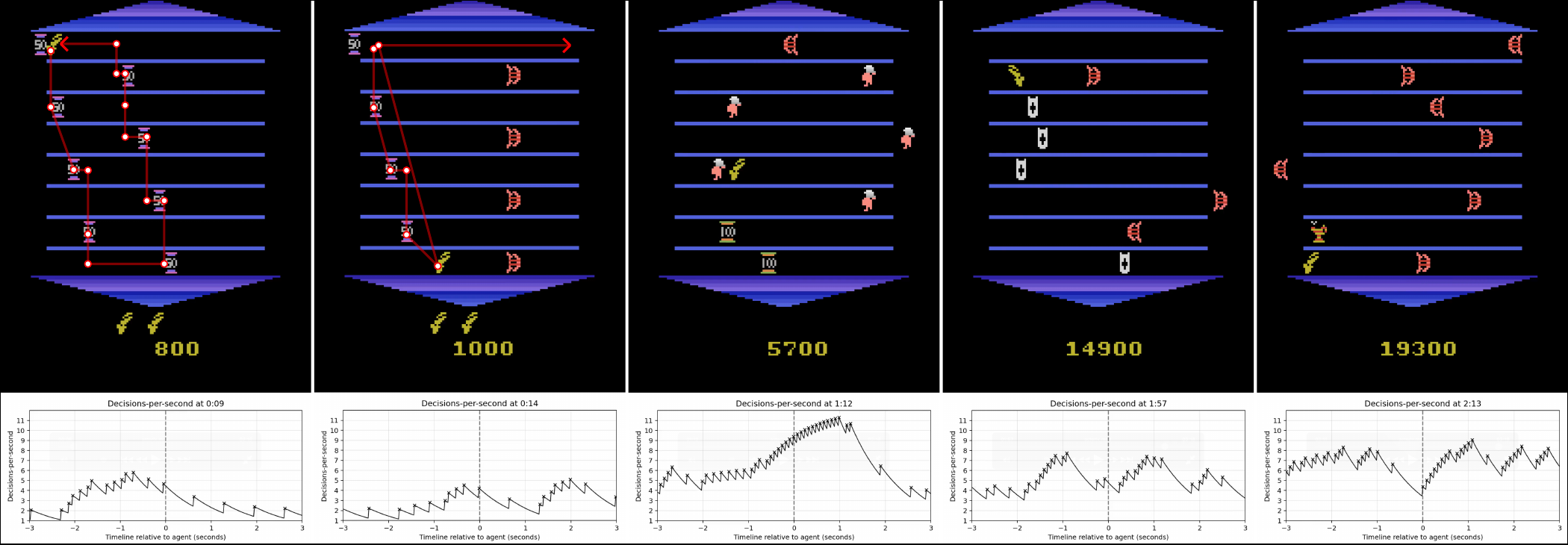}
    \caption{
        A sequence of observations (top) with an associated graph of the exponentially decayed average of decision rate at that moment (bottom) for Compute DQN in Asterix.
        The decision rate grows as the waves get faster and have more collectibles.
        In the first two images, the agent has just finished collecting the objects in a wave, resulting in a spike in decisions followed by a quick decrease awaiting the next wave.
        The remaining images repeat this structure with very high decision rates as the number and speed of objects rises with dips between waves.
    }
    \label{fig:decisions-per-second-asterix-longform}
\end{figure}

Asterix is a game where the agent needs to collect various ancient objects (knights, shields, lamps), and the player needs to avoid the lyre.
As the game progresses, the objects change, and after collecting fifty of one object, the next group of objects appears moving faster than the previous objects.
The agent concentrates compute during dense collectible waves and lowers it between waves (Figure~\ref{fig:decisions-per-second-asterix-longform}), and as objects speed up the decision rate increases proportionally.






\subsection{Agents balance compute costs} \label{sec:agents-respond-to-compute-costs}

\begin{figure}[h]
    \centering
    \noindent\includegraphics[width=\textwidth]{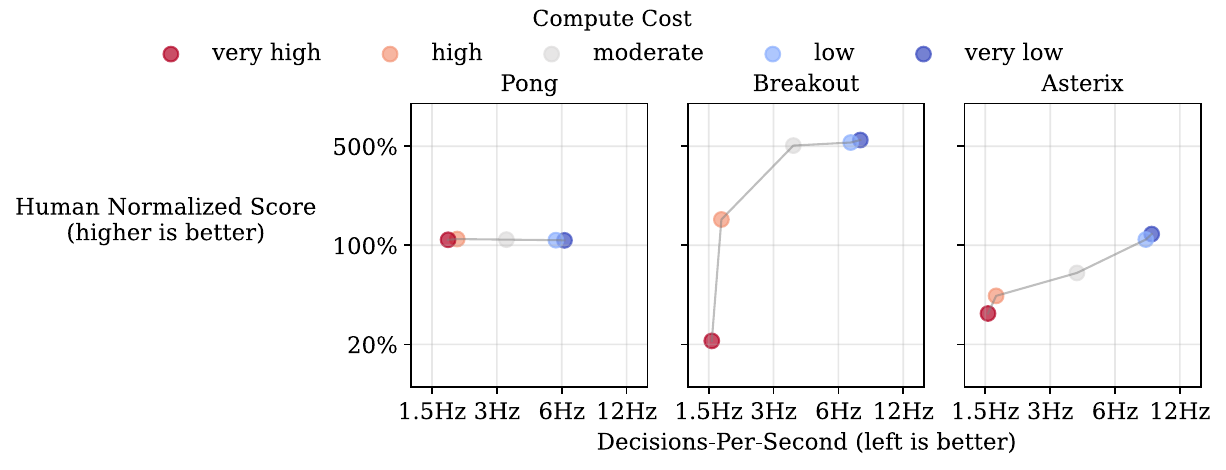}
    \caption{
        Varying the compute cost, $c$, changes the learned decision rate and performance.
        Higher costs yield fewer decisions per-second;
        low costs increase the decision rate.
        Each game shows a performance-compute frontier, showing that agents adapt to their cost parameter.
    }
    \label{fig:responsive-to-compute-costs}
\end{figure}

As a final experiment, we vary the per-decision compute cost, $c$, and observe how these compute costs change the learned policies.
For each cost, we ran 10 seeds for 10 million decisions.
In Figure~\ref{fig:responsive-to-compute-costs}, the moderate setting matches the DQN-derived cost;
we also test higher costs $10c,$ $5c,$ and lower costs $\frac{c}{5}, \frac{c}{10}$.
We see that as the cost increases, policies choose longer options to reduce their decision rate;
as the cost approaches zero, policies operate closer to 12\,Hz.
In Pong, the Pareto tradeoff curve is flat suggesting that a strong policy still exists even at very low decision rates.
We suspect that moderate to low compute costs might provide a small gradient for agents to decrease their decision rate, even though a large reduction in decision rate after 10 million decisions is possible after much more training time (referring back to Figure~\ref{fig:all_games_experience}).
Under higher compute costs, reducing the decision rate is likely prioritized much earlier, which still gives the agent time to learn this compute efficient, yet strong, policy.
In Breakout and Asterix we see a more traditional Pareto curve where performance is sacrificed for compute efficiency under high compute costs.
These results show that agents are respecting the compute-performance tradeoff across high and low cost regimes, while seeing monotonic increases in the use of compute as it becomes cheaper.


\section{Discussion \& related work} \label{sec:discussion}

Several works have considered ideas related to repeating actions.
Some show that action repeats can improve credit assignment~\citep{macro_actions} and sample efficiency~\citep{bellemare_early_atari, mnih_dqn, frameskip_as_a_hyper}, or aid exploration~\citep{dabneyTemporallyextendedEgreedyExploration2020}.
Others allow agents to dynamically select the action repeat length~\citep{aravind_dynamic_frameskip_dqn, sharma_figar, vinyals_alphastar, biedenkapp_temporl}, similar to our approach.
However, these studies are focused on improvements to sample efficiency rather than compute efficiency.
They do not consider a cost for compute, nor how an agent adapts when reducing compute is part of the objective.
A similar approach to ours was considered for continuous-time tasks with a fixed known horizon~\citep{trevenWhenSenseControl2024}, where they reduce their problem to a discrete time RL problem (possibly including an interaction cost).
They investigate problems with relatively short time horizons compared to Atari as their solution involves learning policies with time-remaining as an input.
\citet{holtActiveObservingContinuoustime2023} also study continuous-time control where observations carry a cost, and their agent chooses when it will next observe.

Our work is placed within the long-standing field of conditional computation and meta-reasoning in AI.
In neural networks, early work of  \citet{bengioEstimatingPropagatingGradients2013} and \citet{davisLowrankApproximationsConditional2014} compare several ways to backpropagate through a stochastic neuron, showing that a neural network can learn to conditionally activate costly parts of the model.
Our approach of letting the agent choose its decision rate could be viewed as delaying the rate at which the whole network is activated.
Since, the field has developed~\citep{bengioConditionalComputationNeural2015, bolukbasiAdaptiveNeuralNetworks2017,gravesAdaptiveComputationTime2017, pondernet} showing regularizers that encourage a trade-off between prediction accuracy and compute cost, allowing the neural network to learn when to halt a recurrent computation and output an answer.
The foundational meta-reasoning literature proposes that computations should be treated as actions selected by their expected utility~\citep{russellPrinciplesMetareasoning1991}, with subsequent work developing this idea further~\citep{haySelectingComputationsTheory2012, linMetareasoningPlanningUncertainty2015, callawayLearningSelectComputations2018, buddStopPlannerTime2024}.
These methods operate at a meta-level, where a separate process reasons about an underlying computation;
\citet{buddStopPlannerTime2024}, for example, use deep RL to learn a meta-level policy with the planning cost provided as an input to the agent.
Our agent uses actions and rewards alone, and intentionally so;
we sought to identify whether the RL formulation is sufficient for an agent to reason about its own compute.
And more recently, RL methods could produce intermediate outputs~\citep{hanna_thinking} reminiscent of chain-of-thought reasoning in LLMs~\citep{weiChainofthoughtPromptingElicits2023}, where notions of using compute efficiently are gaining further attention~\citep[e.g.,][]{achilleAIAgentsUniversal2025,raposoMixtureofDepthsDynamicallyAllocating2024}.
These works take a similar approach in that internal actions that modify an agent's computational processes can be treated like any other action.
However, they do not show such agents learning to be more compute efficient or learning to use their compute more deliberately, such as in our experiments on the Arcade Learning Environment.
Finally, the AlphaStar agent~\citep{vinyals_alphastar} imposed a decision-rate limit causing the agent to need to select when it observes the world and acts deliberately, but the focus of their work was to enforce fair play with a human opponent, whereas our work studies an agent's deliberate choice over actions that impact performance on the task and the agent's computation.


We have shown that agents with access to actions that make trade-offs between performance and compute can reason about the impact of these actions on their expected return.
These agents reason by explicitly representing the expected value of selecting a compute-action and how it impacts the return under our explicit compute-performance trade-off.
We have shown that framing a design decision using an option allows agents to reason about a delay in computation;
however, options are not limited to just framing delaying decision-making;
options may be arbitrary sequences of actions;
performed open or closed loop.
Options can also be stochastic or contingent on events~\citep{sutton_options}.
In the longer term, agents could autonomously discover options~\citep{machadoEigenoptionDiscoveryDeep2018, barreto_option_keyboard, harb_options_with_deliberation_cost} with fast execution times that might allow for even lower decision rates.
Adding more available options may increase the complexity of the task for the agent, but with more approximate methods, agents could learn about many options from fragments of experience consistent with executing other options~\citep{suttonIntraoptionLearningTemporally1998}, or plan over them with model-based methods~\citep{suttonDynaIntegratedArchitecture1991, huangOptionZeroPlanningLearned2025}.
Additionally, as people continue to advance agent designs, we expect that agents will similarly be more effective at reasoning about their compute (see additional results with Rainbow in Appendix~\ref{sec:rainbow}), suggesting that algorithmic developments may mitigate any added complexity reasoning about compute may introduce to the RL task.
As the agent designer is concerned with how the agent uses compute and performs on the task, giving the agent actions and feedback that complete this overarching view of the real problem at hand may benefit both the agent and agent designer, as our results have suggested in this paper.

Agents today are typically not long lived~\citep{suttonAlbertaPlanAI2022}: they are trained after which learning is stopped before possible deployment as a fixed policy.
One challenge for agents that continually learn is the big world hypothesis~\citep{javed_bigworld}: the world is large and complex, and designers cannot anticipate how the design of an agent might need to change over time.
If the agent's computational processes are fixed, it may fail to utilize available resources efficiently as the agent experiences new parts of the ``big world''.
Furthermore, a long-lived agent may gain access to more and cheaper compute resources throughout its lifetime;
possibly reflected in lower compute costs.
Such a long-lived agent will need to remain adaptive not just to a constantly changing world, but a constantly availability of resources, or a changing tradeoff of compute costs.


There are many untapped directions for designers to create agents that can reason about their compute.
We hope that future work will investigate providing richer compute-actions beyond controlling the decision rate.
And one would need to be careful that the computational cost of executing these compute-actions is (i) accounted for if not negligible and (ii) much smaller than the computation cost of the agent's typical action selection.

\section{Conclusion}

In this paper we presented Compute DQN, a stepping stone toward agents that can reason about their own computational processes just as they reason about how they can bring about goals in the world.
Our approach aimed to unify the goals of the agent with that of the agent designer, by giving the agent feedback on its computation cost, and providing the agent actions to control how it uses compute by adapting its decision rate.
We showed that with these additions, agents learns interesting and deliberate strategies for using compute that improve with experience and are tailored to their environment and also adapt to their current situation.
We showed performance improvements with the same training budget on 75\% of ALE games, while also showing a 3 times reduction in compute usage.
We believe this work opens up future directions for agents to automatically adapt other aspects of their computational processes, as agents reasoning about them may induce better learned representations and behaviors in the task.
For example, they might be able to adaptively select when they can process observations with a larger or smaller neural network;
how frequently to sample experience from a model;
or adapting batch size or replay ratio~\citep{fedus2020revisiting}.
Furthermore, there is an untapped direction in finding more sample-efficient means to train agents that reason about their compute, such directions could include reward shaping to anneal the cost over time.
In addition, intra-option learning~\citep{suttonIntraoptionLearningTemporally1998}, or methods that generate intra-option tuples of experience~\citep{biedenkapp_temporl}, advancements in model-free and model based methods could help mitigate the additional complexity of the learning problem when giving agents more control over their compute usage.
Agents that adapt could be more energy efficient, making cheaper and faster RL experiments, or agents that make better use of available compute toward greater performance; 
agents could be long-lived by continually adjust compute to needs beyond what can be designed in advance or prescribed by people.

\appendix







\bibliography{references}
\bibliographystyle{rlj}

\beginSupplementaryMaterials

\section{Appendix}

\subsection{DQN and Compute DQN Results}

\begin{figure}[h]
    \centering
    \noindent\includegraphics[width=\textwidth]{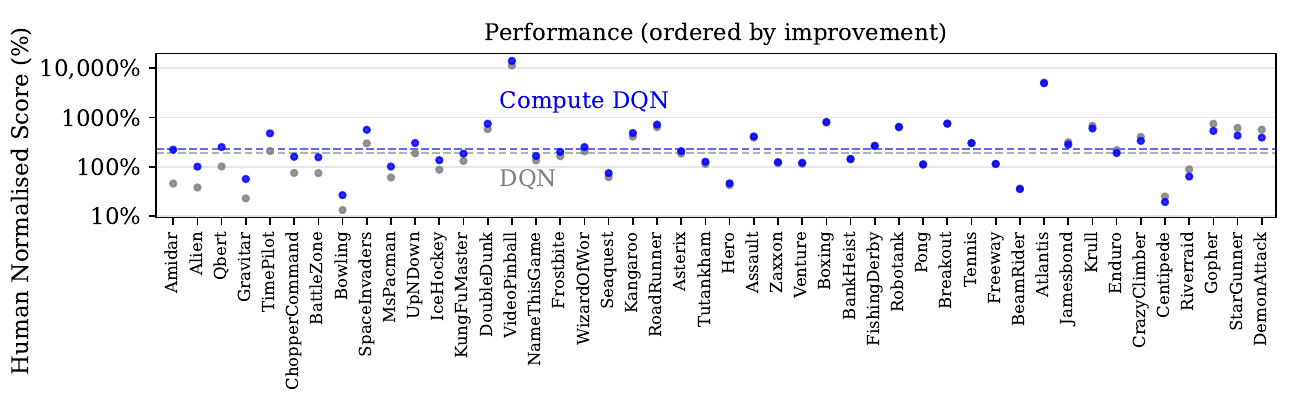}
    \caption{
        We present the human normalized score to validate that (i) DQN and Compute DQN on average perform better than a human (above 100\%), and (ii) the performance loss from Compute DQN (seen on the right-hand side) is marginal relative to the performance benefit (seen on the left).
    }
    \label{fig:arcade_perforamnce_raw}
\end{figure}

\textbf{The raw human normalized score}
is presented to compare both methods across games in Figure~\ref{fig:arcade_perforamnce_raw}.
By computing the geometric average over human-normalized scores, we see that across the ALE, DQN averages 188.58\% HNS, and Compute DQN averages a higher percentage of 232.89\%.
As it is more common to present the number of games improved upon, we report the 75\% of games improved throughout the main text.

\textbf{The number of Decisions, Updates, and Frames } emitted by the emulator is reported in Table~\ref{fig:training_compute_and_frames}.
Compute DQN uses the same amount of training compute as DQN by keeping the decisions and updates equal. 
However, as Compute DQN's mechanism to save compute is to skip frames, this causes the emulator to produce more frames.
We observe that the reduced decision rates seen per-game does not seem to correlate with the performance improvements in that game (see Figure~\ref{fig:overall-decision-rate-per-game}), and so the performance benefit appears to be more largely influenced by the game itself.

\begin{table}
\centering
\caption{
    Compute DQN is trained with the same number of forward passes, decisions, and updates as DQN (40M decisions, 10M updates).
    However, as Compute DQN skips additional frames to reduce its compute cost, the emulator steps more for an agent to be more efficient.
}
\begin{tabular}{l|rrrr}
\toprule
Environment & Decisions (M) & Updates (M) & Emulator Steps (M) \\
\midrule
Alien & 40 & 10 & 1030 \\
Amidar & 40 & 10 & 806 \\
Assault & 40 & 10 & 590 \\
Asterix & 40 & 10 & 571 \\
Atlantis & 40 & 10 & 822 \\
BankHeist & 40 & 10 & 592 \\
BattleZone & 40 & 10 & 594 \\
BeamRider & 40 & 10 & 580 \\
Bowling & 40 & 10 & 637 \\
Boxing & 40 & 10 & 557 \\
Breakout & 40 & 10 & 638 \\
Centipede & 40 & 10 & 815 \\
ChopperCommand & 40 & 10 & 543 \\
CrazyClimber & 40 & 10 & 562 \\
DemonAttack & 40 & 10 & 464 \\
DoubleDunk & 40 & 10 & 262 \\
Enduro & 40 & 10 & 462 \\
FishingDerby & 40 & 10 & 359 \\
Freeway & 40 & 10 & 1341 \\
Frostbite & 40 & 10 & 1008 \\
Gopher & 40 & 10 & 422 \\
Gravitar & 40 & 10 & 440 \\
Hero & 40 & 10 & 859 \\
IceHockey & 40 & 10 & 280 \\
Jamesbond & 40 & 10 & 810 \\
Kangaroo & 40 & 10 & 972 \\
Krull & 40 & 10 & 846 \\
KungFuMaster & 40 & 10 & 566 \\
MsPacman & 40 & 10 & 1034 \\
NameThisGame & 40 & 10 & 557 \\
Pong & 40 & 10 & 710 \\
Qbert & 40 & 10 & 1018 \\
Riverraid & 40 & 10 & 748 \\
RoadRunner & 40 & 10 & 710 \\
Robotank & 40 & 10 & 351 \\
Seaquest & 40 & 10 & 768 \\
SpaceInvaders & 40 & 10 & 942 \\
StarGunner & 40 & 10 & 530 \\
Tennis & 40 & 10 & 452 \\
TimePilot & 40 & 10 & 730 \\
Tutankham & 40 & 10 & 904 \\
UpNDown & 40 & 10 & 704 \\
Venture & 40 & 10 & 335 \\
VideoPinball & 40 & 10 & 814 \\
WizardOfWor & 40 & 10 & 410 \\
Zaxxon & 40 & 10 & 337 \\
\bottomrule
\end{tabular}
\label{fig:training_compute_and_frames}

\end{table}

\newpage

\subsubsection{Training DQN using the average decision rate learned by Compute DQN} \label{sec:dqn_using_learned_avg_decision_rate}

\begin{figure}[h]
    \centering
    \noindent\includegraphics[width=\textwidth]{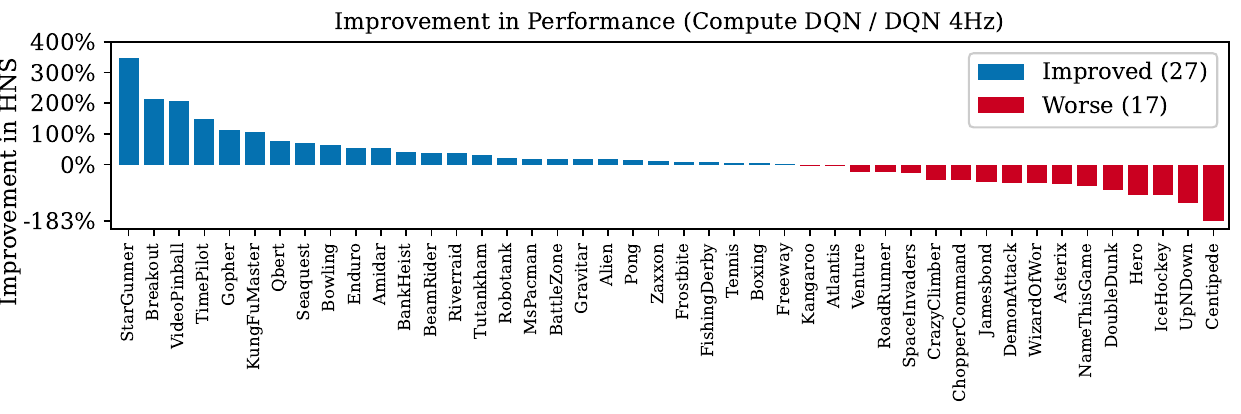}
    \caption{
        Improvement of Compute DQN over DQN (at 4\,Hz) in human normalized score (HNS) across 44 Atari games. 
        Compute DQN achieves higher HNS in 61\% of games. 
        StarGunner shows the largest gain of 349\%, while the largest loss of 183\% below DQN.
    }
    \label{fig:arcade_perforamnce_improvement_4hz}
\end{figure}

We train DQN with options of a fixed duration of 15 frames to determine whether there is a performance improvement from operating at this fixed rate, as opposed to giving the agent actions and feedback to reason about compute for themselves.
We chose 15 frames, or 4\,Hz, over exactly matching 3.6\,Hz to provide DQN more access to compute during play on average than what is learned by Option DQN.
If DQN performs higher at 4\,Hz, this indicates that there exist ways to play each game such that DQN could learn to play very efficiently at a much lower decision rate.
Our investigation here is primarily focused on analyzing the change in performance, and so, we see in Figure~\ref{fig:arcade_perforamnce_improvement_4hz} that Compute DQN can still perform better on 61\% of games.
This suggests that Compute DQN benefits from having access to the additional options to adapt its decision rate on a per-game basis and to each moment within a game. 

\vspace{-0.5em}

We present the human normalized score to compare both methods across games in Figure~\ref{fig:arcade_perforamnce_improvement_4hz}.
By computing the geometric average over human normalised scores, we see that DQN (4\,Hz) averages 202\% HNS across the ALE, and Compute DQN averages a higher percentage of 225\%.
As it is more common to present the number of games improved upon, we see that Compute DQN is better on 61\% of games while achieving the same compute efficiency as DQN fixed to 4\,Hz.

\begin{figure}[h]
    \centering
    \noindent\includegraphics[width=\textwidth]{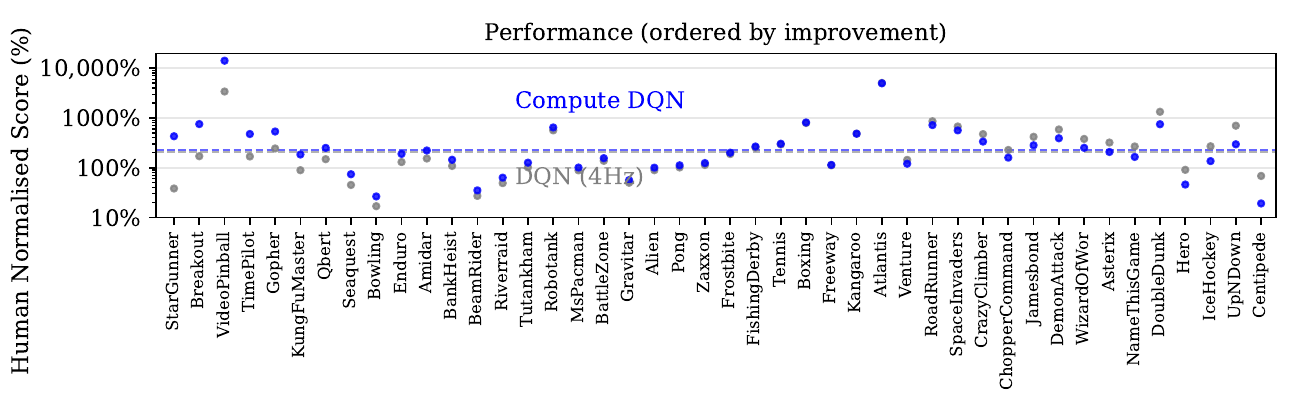}
    \caption{
        We present the human normalized score to validate that (i) DQN (4\,Hz) and Compute DQN on average perform better than a human (above 100\%).
        The performance gains for StarGunner and Breakout (both left), and the performance loss on Centipede are large (right). 
        On average, we see that giving agents control over their compute still results in a performance and efficiency improvement averaged across the suite.
    }
    \label{fig:arcade_raw_perforamnce_improvement_4hz}
\end{figure}

\newpage
\subsubsection{Rainbow} \label{sec:rainbow}

\begin{figure}[h]
    \centering
    \noindent\includegraphics[width=\textwidth]{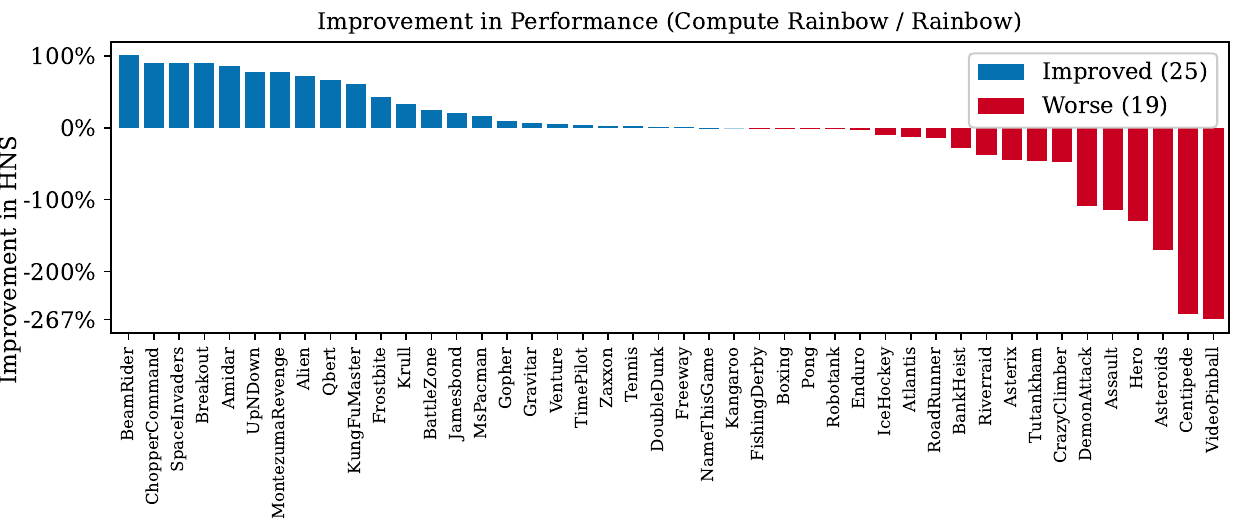}
    \caption{
        Improvement of Compute Rainbow over Rainbow.
    }
    \label{fig:rainbow_arcade_perforamnce_improvement}
\end{figure}

\begin{figure}[h]
    \centering
    \noindent\includegraphics[width=\textwidth]{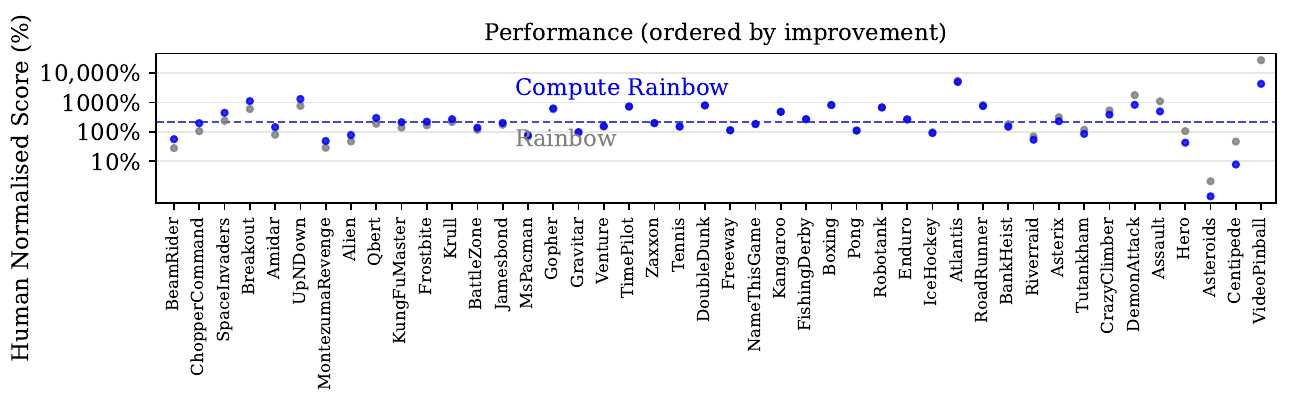}
    \caption{
        Performance of Compute Rainbow and Rainbow.
    }
    \label{fig:rainbow_raw_arcade_performance}
\end{figure}

In this section we investigate whether improvements in performance and efficiency are improved from improvements in RL solution methods.
Figure~\ref{fig:rainbow_arcade_perforamnce_improvement} and Figure~\ref{fig:rainbow_raw_arcade_performance} show Compute Rainbow is better in just over half (57\%) of the games than Rainbow. 
Surprisingly, we see a much lower decision rate on average is learned across games, where Compute Rainbow averages just above 2\,Hz, or an efficiency improvement of approximately 6 times.
This may be explained by Rainbow having improvements in credit assignment from n-step returns, an improved method of exploration, using prioritized experience replay. 

\begin{figure}[h]
    \centering
    \noindent\includegraphics[width=\textwidth]{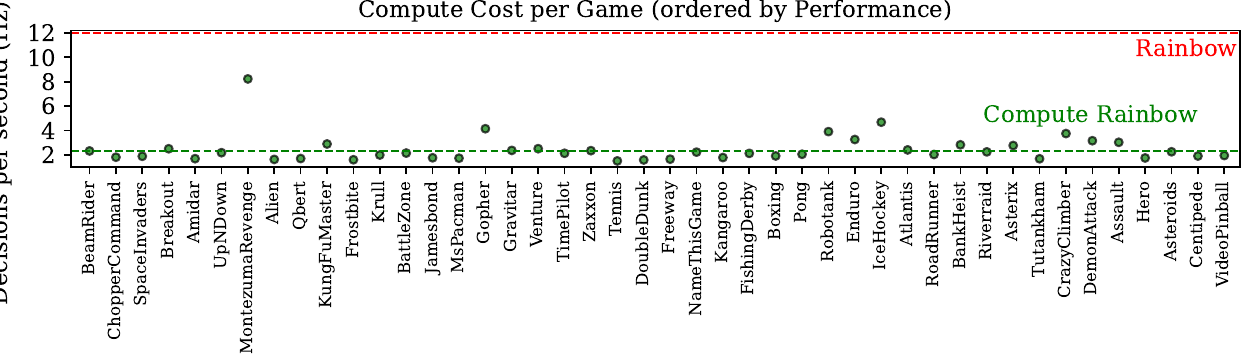}
    \caption{
        Compute cost of Compute Rainbow and Rainbow.
    }
    \label{fig:rainbow_compute_cost}
\end{figure}

\end{document}